\documentclass{article}

\usepackage{natbib}

\usepackage[final]{nips_2018}

\usepackage{graphicx}
\usepackage{amsmath,latexsym,amssymb,amsthm,bbm,array,multirow}
\usepackage{soul}

\usepackage[utf8]{inputenc} 
\usepackage[T1]{fontenc}    
\usepackage{hyperref}       
\usepackage{url}            
\usepackage{booktabs}       
\usepackage{amsfonts}       
\usepackage{nicefrac}       
\usepackage{microtype}      

\usepackage{xargs}
\usepackage[colorinlistoftodos,prependcaption,textsize=tiny]{todonotes}
\newcommandx{\chaofan}[2][1=]{\todo[linecolor=blue,backgroundcolor=blue!25,bordercolor=blue,#1]{Chaofan:#2}}
\newcommandx{\kangcheng}[2][1=]{\todo[linecolor=blue,backgroundcolor=blue!25,bordercolor=blue,#1]{Kangcheng:#2}}
\newcommandx{\cynthia}[2][1=]{\todo[linecolor=blue,backgroundcolor=blue!25,bordercolor=blue,#1]{Cynthia:#2}}
\newcommandx{\yaron}[2][1=]{\todo[linecolor=blue,backgroundcolor=blue!25,bordercolor=blue,#1]{Yaron:#2}}
\newcommandx{\sijia}[2][1=]{\todo[linecolor=blue,backgroundcolor=blue!25,bordercolor=blue,#1]{Sijia:#2}}
\newcommandx{\tong}[2][1=]{\todo[linecolor=blue,backgroundcolor=blue!25,bordercolor=blue,#1]{Tong:#2}}

\title{An Interpretable Model with Globally Consistent Explanations for Credit Risk}

\author{
  Chaofan Chen$^{1}$, Kangcheng Lin$^{2}$, Cynthia Rudin$^{1,2,3}$, Yaron Shaposhnik$^{4}$, Sijia Wang$^{3}$, Tong Wang$^{5}$ \\
  $^{1}$ Department of Computer Science,
  Duke University \\
  $^{2}$ Department of Statistical Science, Duke University \\
  $^{3}$ Department of Electrical and Computer Engineering, Duke University \\
  $^{4}$ Simon Business School, University of Rochester \\
  $^{5}$ Department of Management Sciences, Henry B. Tippie College of Business, The University of Iowa  \\
}

\begin{document}

\maketitle

\begin{abstract}

 We propose a possible solution to a public challenge posed by the Fair Isaac Corporation (FICO), which is to provide an explainable model for credit risk assessment. Rather than present a black box model and explain it afterwards, we provide a globally interpretable model that is as accurate as other neural networks. Our "two-layer additive risk model" is decomposable into subscales, where each node in the second layer represents a meaningful subscale, and all of the nonlinearities are transparent. We provide three types of explanations that are simpler than, but consistent with, the global model. One of these explanation methods involves solving a minimum set cover problem to find high-support globally-consistent explanations. We present a new online visualization tool to allow users to explore the global model and its explanations.\footnote{All authors contributed equally to this work. Authors are listed alphabetically.}
\end{abstract}

\section{Introduction}
In 2018, the Fair Isaac Corporation (FICO) proposed a challenge to data science researchers: present an explainable model for the risk of defaulting on a loan. FICO provided a dataset for the challenge, and asked researchers to provide explanations for the global model, as well as local explanations (explanations for a given prediction). They also requested that the global model respect monotonicity constraints (increasing or decreasing) on several of the variables.

In responding to this challenge, we considered the specific aspects of the dataset in our modeling approaches: 
the FICO data are balanced between the two classes, most of the features are real-valued, and most importantly, each of the 23 features is itself interpretable. Because the features already come with a good representation, the algorithm does not need to construct the representation. Generally, for data having this particular property,  perhaps with a small amount of feature engineering, most machine learning algorithms tend to have almost the same performance, including algorithms that produce globally interpretable models. Thus, we aimed to create a model that was fully and globally interpretable, 
rather than to construct a black box.

We call our globally interpretable model a \textit{two-layer additive risk model}. It was designed to resemble traditional subscale models, where the features are partitioned into meaningful subgroups, and the subgroup scores are later combined into a global model. Traditional subscale models are generally interpretable because they are decomposable into meaningful components, and because these models are usually linear with coefficients whose sign is positive for risk factors. Our model preserves these classical elements (decomposable, uses linear modeling, positive coefficients for risk factors), but inserts (interpretable) nonlinearies in several places to make the model more flexible and accurate. In particular, the algorithm transforms the original features into piecewise constant functions that monotonically increase (or decrease) if we constrain them to do so. Combinations of these piecewise constant functions form the subscales in the second layer of the network. The subscales are fed through a sigmoid nonlinearity, which has the effect of adding more perceptron-like flexibility to the model, but also makes the subscale scores more meaningful as their own ``mini-models;'' each subscale produces its own probability of defaulting on a loan. The subscales are combined linearly and sent through a sigmoid function to produce the final probability of defaulting on a loan. 

While working on this challenge, it is important to note that the guidelines asked for an explanation for each class of the global model, such as variable importance information. It did not necessarily ask for a model that is globally interpretable. Perhaps it could benefit FICO to have a global model that is \textit{not} interpretable (a ``secret sauce'')? It is not clear that even if there exists a globally interpretable model, such as the one we found, that it would be desirable for FICO to release it. Thus, we are not certain that we responded as much to FICO's needs as we did to its customers' needs. 

The issue raised above, about explainability of a black box model, versus providing a globally interpretable model, is important. With the new General Data Protection Regulation (GDPR) \citep{regulation2016general} regulations such as ``right to an explanation,'' there could still be little incentive for companies to provide anything more than a modestly local explanation, even if a globally interpretable model existed. However,  explanations can be problematic for several reasons. First, unless the global model is uniformly equal to the local model, the explanations will be sometimes incorrect, which makes it difficult to trust either the explanations or the global model itself. Even if an explanation gives the same prediction as the global model, it could be inconsistent with the global model's actual calculations (i.e., low fidelity \citep{guidotti2018survey}), or it might provide reasons that may be true for some cases but not others. For instance, a reason of ``too many accounts open'' may be used to deny someone a loan even if there is another person with the same number of open accounts who was offered a loan by the same global model. Explanations could also be correct but misleading, offering reasons that are true but incomplete -- missing key information. All of these problems with explanation-for-black-box methods are reasons that consumers would benefit from globally interpretable models. Again, however, we fully recognize that creating a globally interpretable model is not usually desirable from a business perspective.\footnote{Companies such as DivePlane are now grappling with this issue, where they do not, as of this writing, release code or demonstrations of their models, which are claimed not to be black boxes.}

Attempts to create globally interpretable models for financial applications use mainly standard machine learning approaches (e.g., decision trees and support vector machines are used for bank direct marketing \citep{moro2011data, moro2014data}), which cannot accommodate FICO's monotonicity constraints. Some work finds optimal rules \citep{chen2018optimization}, but rules are not natural for datasets with many real-valued features, like FICO's data. Additive models are natural for real-valued features and can easily preserve monotonicity. 


Even though our global model is interpretable and thus can be explained on its own, we can also produce optional local ``explanations,'' which now simply become \textit{summaries} of general trends in the global model. These are summaries rather than explanations (or \textit{summary-explanations}) in that they do not aim to reproduce the global model, only to show patterns in its predictions. Our summary-explanation method (called \texttt{SetCoverExplanation}) is a model-agnostic explanation algorithm \citep{ShaposhnikRu18}. \texttt{SetCoverExplanation} solves a minimal set cover optimization problem \citep{feige1998threshold,feige1996threshold} to generate conjunctive rules that are consistent with all training cases.  

To create an explanation of an individual prediction, our interactive display first highlights the factors that contribute most heavily to the final prediction of the global model. Second, we show patterns produced by \texttt{SetCoverExplanation}.
Third, we provide case-based explanations. Our case-based reasoning method finds cases that are similar on important features to any current case that the user inputs.

We created an interactive display that shows the full computation of the model from beginning to end, without hiding any nonlinearities or computations from the user. Factors are colored according to their contribution to the global model. The form of our global model lends itself naturally to variable importance analysis, and understanding monotonicity constraints, through the visualization.

The novel elements of the work are (i) the form of the two-layer additive risk model, which lends naturally to sparsity, decomposibility, visualization, case-based reasoning, feature importance, and monotonicity constraints, (ii) the interactive visualization tool for the model, (iii) the use of the \texttt{SetCoverExplanation} algorithm for high-support local conjunctive explanations, and (iv) the application to finance, indicating that black boxes may not be necessary in the case of credit-risk assessment.



\section{Two-Layer Additive Risk Model and its Visualization}
\label{gen_inst}
We work with a dataset $\{(\mathbf{x}_i, y_i)\}_{i=1}^N$, where $\mathbf{x}_i \in \mathbbm{R}^P$ is a vector of features, where the categorical features are binarized. The labels are indicators of defaulting on a loan: $y_i \in \{0,1\}$. Let $\mathcal{P}$ represent the set of features and $|\mathcal{P}| = P$.
We present the structure of our two-layer additive risk model (ARM).  While in general, neural networks are hard to comprehend even with sparsity regularization applied, our model has carefully designed sparsity and monotonicity constraints that make the calculations easier to comprehend. Moreover, our model did not require quadratic terms, as \citep{Caruana13} did.

\begin{figure}
        \centering
        \includegraphics[width=.20\linewidth]{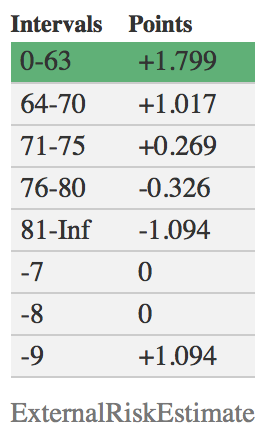}
        \caption{The ``External Risk Estimate" Subscale: sum of thresholds, as a piecewise constant function, written as a scoring system. Because the value of the external risk estimate is less than 63, the person receives 1.799 points.}
        \label{fig:ExternalRiskEstimate}
    \end{figure}

First, to ensure monotonicity of the model with respect to any given features, we used step functions as our initial transformations of the features, and constrained the coefficients of each feature's step functions to be non-negative. For instance, for a monotonically decreasing feature $x_{\cdot,p}$, the following features could be created: $b_{p,1}(x_{\cdot,p})=\mathbbm{1}[x_{\cdot,p} < 10]$, $b_{p,2}(x_{\cdot,p})=\mathbbm{1}[x_{\cdot,p} < 50]$, $b_{p,3}(x_{\cdot,p})=\mathbbm{1}[x_{\cdot,p} < 75]$, and $b_{p,0}(x_{\cdot,p})=\mathbbm{1}[x_{\cdot,p} \text{ is not missing}].$\footnote{We handle missing values also by creating binary indicator features for missingness.} 
    Note that all of these 
    features use one-sided intervals. This choice was made because convex combinations of these step functions yield \textit{monotonic} piecewise linear functions. Specifically,
    by enforcing the constraints that the coefficients for $b_{p,1}$, $b_{p,2}$, and $b_{p,3}$ must be non-negative, we shall guarantee a monotonically decreasing relationship between the original continuous feature $x_{\cdot,p}$ and the subscale's predicted probability of default. Once the model is learned, the sum of the one-sided intervals becomes a piecewise constant function. For instance,
    \[
    f_p(x_{\cdot,p}) = \beta_{p,1} b_{p,1}(x_{\cdot,p}) + \beta_{p,2} b_{p,2}(x_{\cdot,p}) + \beta_{p,3} b_{p,3}(x_{\cdot,p}) + \beta_{p,0} b_{p,0}(x_{\cdot,p})
    \]
    can be equivalently written as
    \begin{eqnarray*}
    f_p(x_{\cdot,p}) &=& (\beta_{p,1}+\beta_{p,2}+\beta_{p,3}+\beta_{p,0})\mathbbm{1}[x_{\cdot,p}<10] + (\beta_{p,2}+\beta_{p,3}+\beta_{p,0})\mathbbm{1}[10 \leq x_{\cdot,p} < 50] \\
    &&+ (\beta_{p,3}+\beta_{p,0})\mathbbm{1}[50 \leq x_{\cdot,p} < 75] + \beta_{p,0}\mathbbm{1}[75 \leq x_{\cdot,p} ],
    \end{eqnarray*}
    which can be displayed as a traditional scoring system \citep{ustun2016supersparse,mdcalc}. We show a transformation like this in Figure \ref{fig:ExternalRiskEstimate} for the ExternalRiskEstimate subscale. 
    \begin{figure}
        \centering
        \includegraphics[width=.3\linewidth]{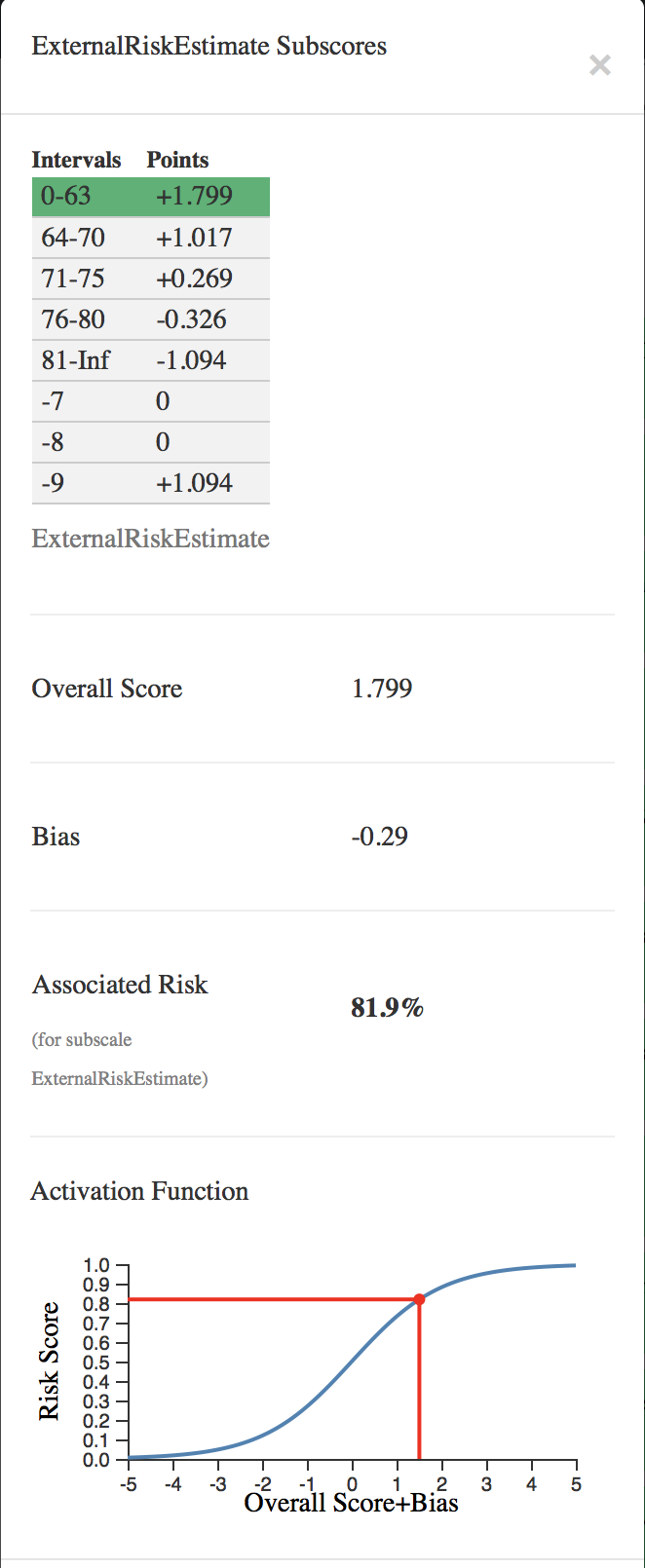}\;\;
         \includegraphics[width=.61\linewidth]{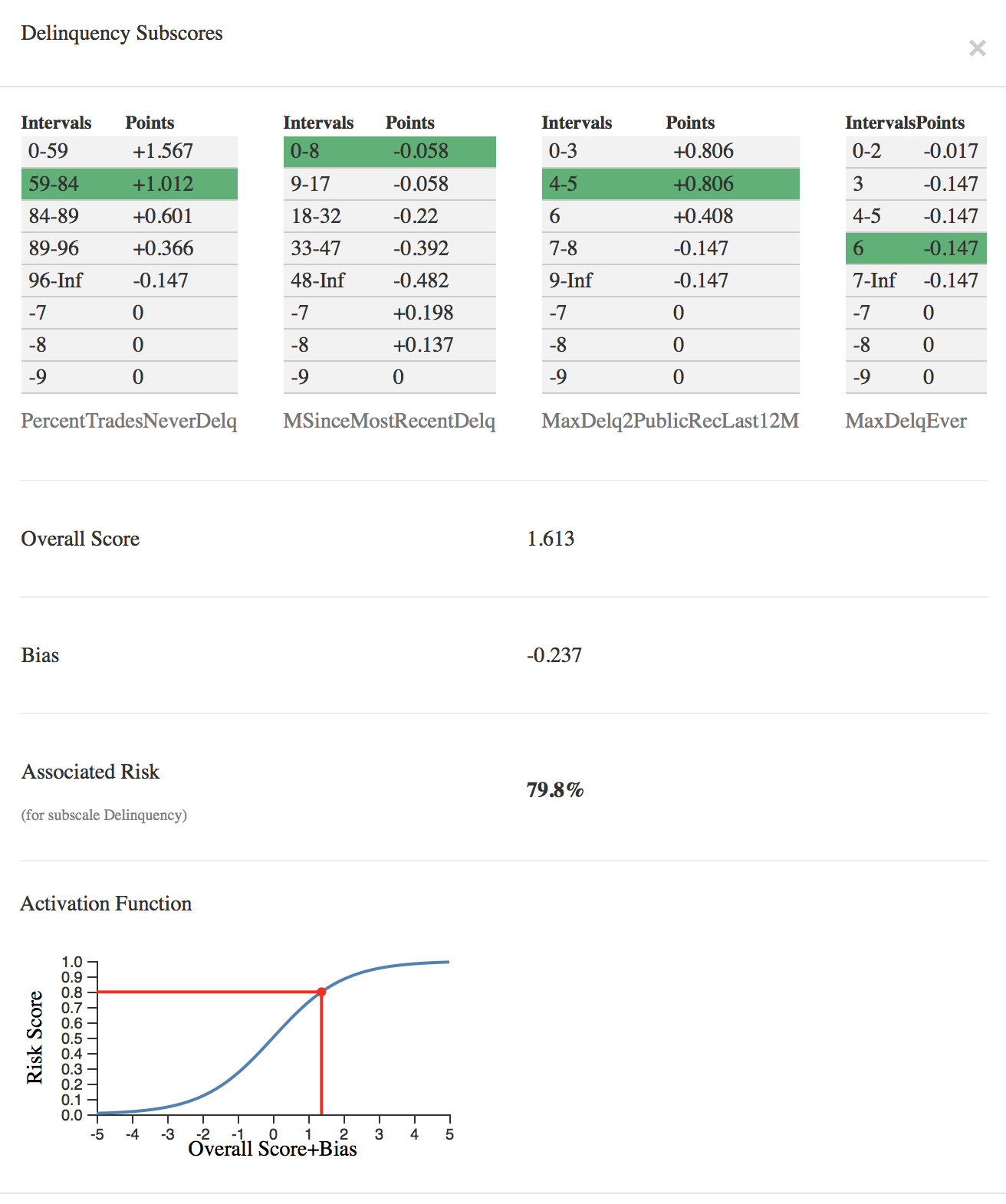}
        \caption{The ``External Risk Estimate" and ``Delinquency'' Subscales.  \textit{Left}: The transformation of points into a risk estimate, which is 81.9\% for this person. The risk estimate for the subscale is meaningful as its own ``mini-model'' of risk, based only on the ExternalRiskEstimate feature. \textit{Right}: A different subscale,  which uses multiple features. }
        \label{fig:subscales}
    \end{figure}
    
    If coefficients $\beta_{p,1}$, $\beta_{p,2}$, and $\beta_{p,3}$, are nonnegative, the function $f_p$ is nonincreasing. 
    If instead we would like to constrain $f_p$ to be monotonically increasing, we reverse the above one-sided inequalities $<$ into $>$. Of course, if a feature $x_p$ has no desired monotonicity, we drop non-negativity constraints.

Using domain knowledge obtained from the data description, we partitioned the features $\mathcal{P}$ into different sets for the subscales, inducing sparsity in the first layer of the network. Each subset of features is sent to one node which computes a subscale. Denote the feature subsets as 
\begin{equation*}
    \mathcal{P} = \cup_{k=1}^K\mathcal{P}^{[k]},
\end{equation*}
where subset of features $\mathcal{P}^{[k]}$ is sent to a subscale. There are between 1 and 4 of the original features combined per subscale, yielding a total of 10 subscales to represent the original 23 features.
Each subscale can be interpreted as a miniature model 
    for predicting the probability of failure to repay a loan, using only the features designated for the subscale. 
    The output of the subscale is a probability, denoted by $r^{[k]}$ for subscale $k$, which is simply a sigmoid transformation of the score, 
    \begin{equation*}
        r^{[k]}(\mathbf{x}) = \sigma\left(\sum_{p\in \mathcal{P}^{[k]}} f_p(x_{\cdot,p})\right) = \sigma\left(\sum_{p\in \mathcal{P}^{[k]}} \sum_{l=0}^{L_p} \beta_{p,l} b_{p,l}(x_{\cdot,p})\right),
    \end{equation*}
where $\sigma(\cdot)$ represents a sigmoid function and $L_p$ is the number of binary features created for feature $p$ (not counting the special indicator for non-missing values).

Finally, the subscale results are linearly combined and again nonlinearly transformed into a final probability of failure to repay a loan. The contribution of each subscale to the final prediction can be easily observed by its weighted output.

\begin{figure}[t]
    \centering
    \includegraphics[width=.65\linewidth]{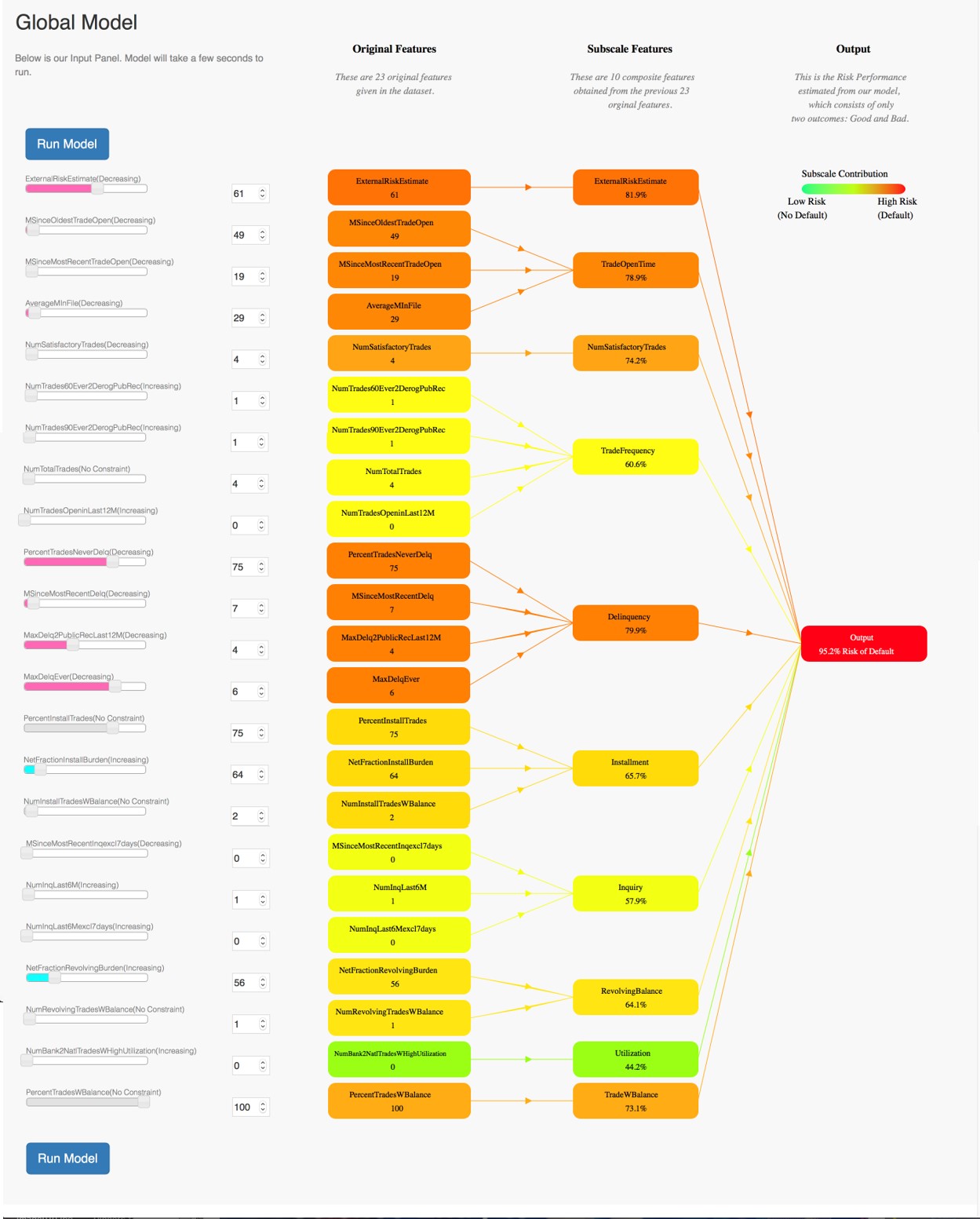}
    \includegraphics[width=.34\linewidth]{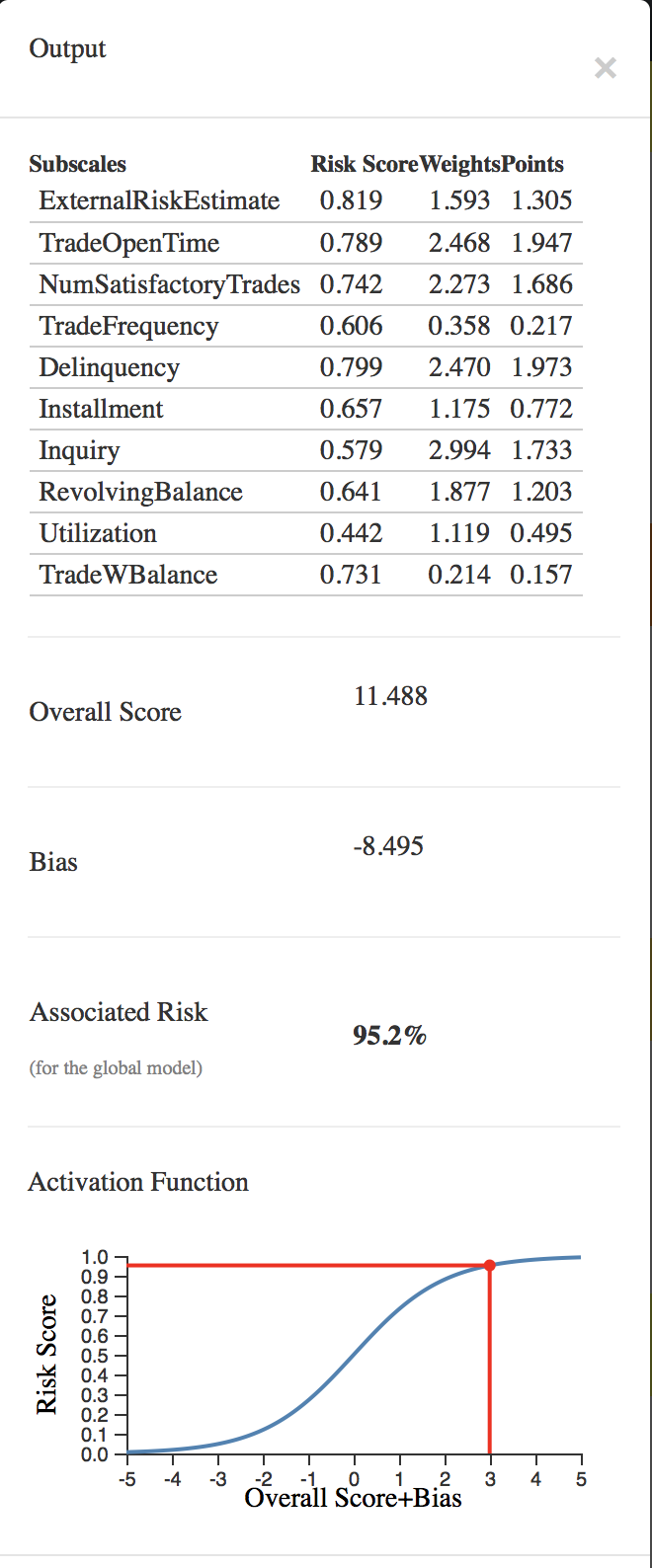}
    \caption{\textit{Left}: Snapshot of visualization tool showing the global model. Colors indicate contribution to the final score. The 23 feature values are entered on the left. \textit{Right}: Final combined score pop-up.}
    \label{fig:FullModel}
\end{figure}

\begin{figure}[t]
\centering
\includegraphics[width=0.547\linewidth]{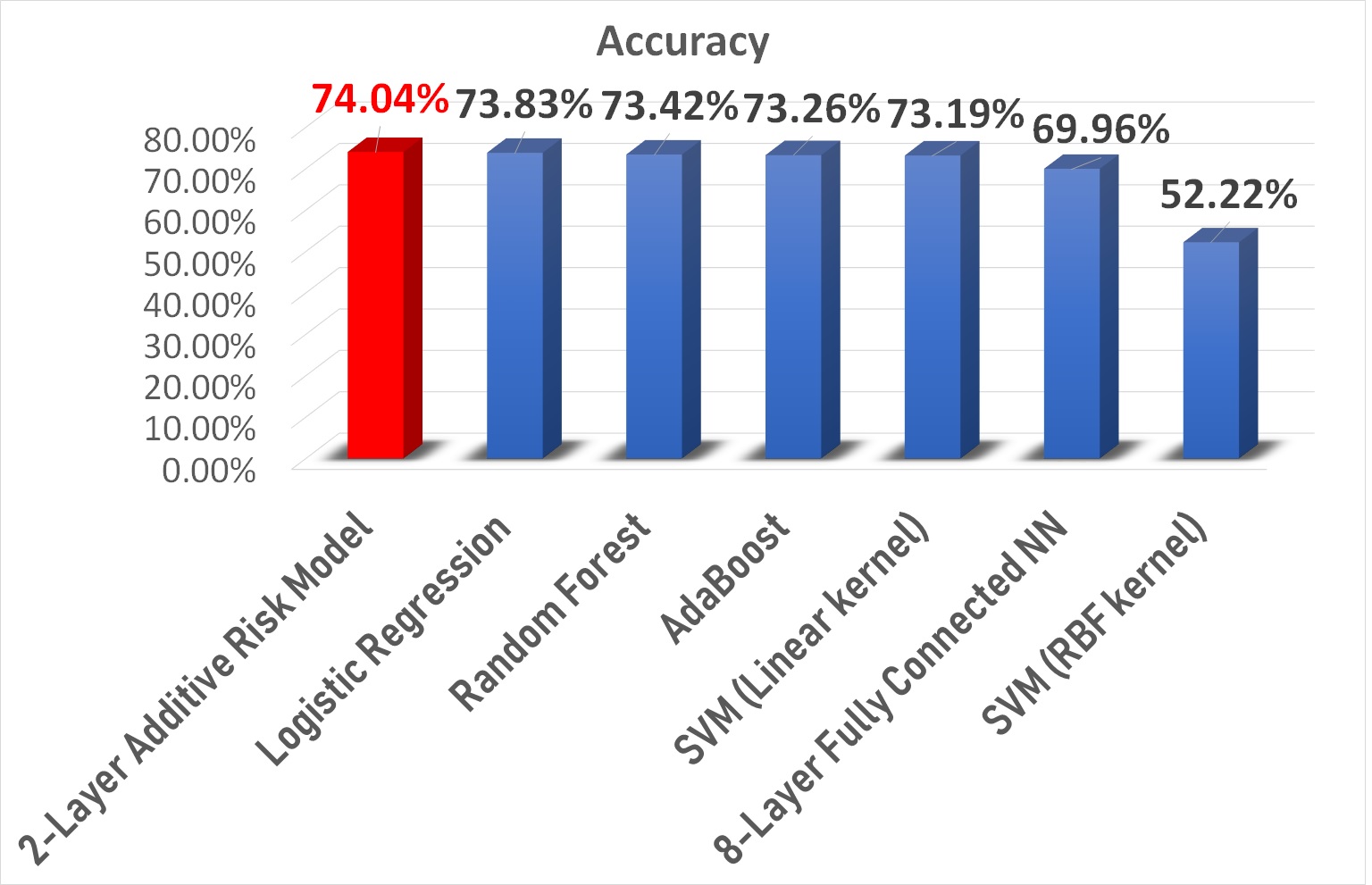}
\caption{The accuracy of our model compared to other common machine learning models.}
\label{fig:AccuracyPlot}
\end{figure}

The simplest way to train coefficients 
$\boldsymbol{\beta}^{[k]} = \{\beta_{p,l}\text{ for } p \in \mathcal{P}^{[k]}, l \in [0, L_p] \}$
is to treat each $r^{[k]}$ as an independent classification model with the $\{y_i\}_{i=1}^N$ being the target variables, using regularization (e.g., $\ell_2$) to prevent overfitting, and positivity constraints on the thresholds to enforce monotonicity. A slightly more complicated way to train is to optimize for a combination of accuracy for the subscales and accuracy for the global model. On our data, these two methods tended to produce almost identical results. An image of the full model is shown in Figure \ref{fig:FullModel}, where the colors indicate the final contribution to the combined score. Red indicates more likely to default on the loan. The 23 feature values can be entered on the left, and clicking on any of the 10 subscales (in the second colored layer) reveals a pop-up window with the calculation, as shown in Figure \ref{fig:subscales} for two subscales. The final combination of features is shown in Figure \ref{fig:FullModel} (right panel). Figure \ref{fig:AccuracyPlot} shows that our global model does not lose accuracy over other machine learning techniques, despite being constrained to be interpretable. The accuracy results were obtained by averaging test accuracy figures of five random $80\%:20\%$ training-test splits. Our final model was trained on the entire dataset.

\subsection*{Variable Importance}
Our two-layer additive risk model comes naturally with a way to identify a list of factors that contribute most heavily to the final prediction.
Table~\ref{tab:ImportantFeatures} shows a list of four factors that are important for predicting observation ``Demo 1''  to have $95.2\%$ risk of default (i.e., bad risk performance). 

\begin{table}[]
    \centering
    \caption{An example of the most important contributing factors learned for observation ``Demo 1'' }
    \begin{tabular}{c|l}
\toprule
         & \multicolumn{1}{c}{\textbf{Most important contributing factors}} \\ \hline 
      1   & MaxDelq2PublicRecLast12M is 6 or less (from the most important subscale, Delinquency)\\ 
      2 & PercentTradesNeverDelq is 95 or less (from the most important subscale, Delinquency)\\ 
      3 & AverageMInFile is 48 or less (from the second most important subscale, TradeOpenTime) \\ 
      4 & AverageMInFile is 69 or less (from the second most important subscale, TradeOpenTime) \\
         \bottomrule
    \end{tabular}
    
    \label{tab:ImportantFeatures}
\end{table}
To identify the factors, we first identify the most important two subscales and then the most important factors within each subscale.
The importance of each subscale in the final model is determined by its weighted score, which is the product of the subscale's output and its coefficient -- the larger the product, the larger the contribution of the term in the final risk. 

For example, for ``Demo 1'', the two most important subscales are Delinquency (with points of 1.973) and TradeOpenTime (with points of 1.947). Then, within each of the two subscales, we find two factors that contribute the most to that particular subscale's risk score. 
The two most important factors for each subscale are determined likewise by the product of the coefficient of each binary feature and the value of the binary feature itself. 
We finally output those binary features and their corresponding values as the most important contributing factors to the prediction made by our model.

The factors are grouped by subscales and are displayed in decreasing order of importance (within each important subscale) to the global model's predictions.

\section{Consistent rule-based explanations with \texttt{SetCoverExplanation}}

As noted earlier, in addition to predicting risk using a globally interpretable model, we generate consistent rules that summarize broad patterns of the classifier with respect to the data. They do not explain the global model's computations; instead they provide useful patterns, which has been a popular form of explanation in the state-of-the-art literature on model explanations \citep{lakkaraju2017interpretable,guidotti2018local,ribeiro2018anchors}. As an example, consider Observation 6 in the FICO dataset, for which the global model predicts a high risk of default. \texttt{SetCoverExplanation}  returns the following rule-based summary-explanation that includes Observation 6:

\noindent\fbox{%
    \parbox{\textwidth}{%
        For \textit{all} 700 people where: 
        \begin{itemize}
            \item ExternalRiskEstimate$\leq 63$ , and
            \item  NetFractionRevolvingBurden$\geq 73$, 
        \end{itemize}
         the global model predicts a high risk of default.
    }%
}

\texttt{SetCoverExplanation} asserts that our global model predicts high-risk for \textit{all} of the 700 previous cases that satisfy these rules. Therefore these rules are \textit{globally consistent}. In contrast, explanations (from other methods) that are not consistent may hold for one customer but not for another, which could eventually jeopardize trust.  

In what follows, we formalize the discussion on consistent rules and put it into concrete mathematical terms. 
After defining rules in Section~\ref{sub:rules_notation}, we address aspects of optimization in Section~\ref{sub:rules_optimization}. In Section~\ref{sub:consistent-cases} we describe how to use consistent rules to identify similar cases for case-based explanations.

\label{headings}
\subsection{Notation and definitions}\label{sub:rules_notation}

Consider a $P$-dimensional binary data set $\{(\mathbf{x}_i, y_i)\}_{i=1}^N$, that is, $x_{i,p}\in\{0,1\}$ for every $i$ and $p$. Let $h^M:\{0,1\}^P\rightarrow \{0,1\}$ denote a classifier that was trained using the dataset, and let $\textbf{y}^M$ denote the vector of labels generated by the model $h^M$ (the super-script $M$ stands for {model}). 

Assume $\mathcal{P}'\subseteq \mathcal{P}$ is a subset of features and ${y}\in\{0,1\}$ is a label. The \textit{rule} $\mathcal{P}'\Rightarrow {y}$ describes the following binary function/classifier:
\[ h^{\mathcal{P}'\Rightarrow {y}}\left(\textbf{x}\right) =
  \begin{cases}
    {y}       & \quad \text{if } \left(\prod_{p\in \mathcal{P}'} x_{\cdot,p}\right)=1 \\
    1-{y}     & \quad \text{otherwise.}
  \end{cases}
\]
That is, for a given observation $\textbf{x}$, the rule $\mathcal{P}'\rightarrow {y}$ predicts ${y}$ based on the projection of the observation onto the subspace of features $\mathcal{P}'$; specifically, by applying a logical AND operator on the subset of features in $\mathcal{P}'$. 

Let $\textbf{x}_e, y_e$ denote an observation and the respective model prediction that we wish to create a rule for. We say that the rule $\mathcal{P}'\Rightarrow y_e$ provides a \textit{consistent summary-explanation} for $\textbf{x}_e, y_e$ if the following conditions are met: 
\begin{enumerate}
\item (Relevance) $x_{e,p}=1$ for every $p\in \mathcal{P}'$.
\item (Consistency) For every observation $\textbf{x}_i$ for which $x_{i,p}=1$ for all $p\in \mathcal{P}'$, it must also hold that $y^M_i=y_e$.
\end{enumerate}
The second condition establishes consistency by enforcing all observations in the dataset to agree with the rule, in the sense that all observations $i$ for which the binary variables in $\mathcal{P}'$ are true are similarly labeled by the global model $M$ as $y_i^M=y_e$. 

We measure the quality of a rule $\mathcal{P}'\rightarrow {y}$ using two criteria:
\begin{itemize}
\item \textit{Sparsity} -- the cardinally of $\mathcal{P}'$, that is, $|\mathcal{P}'|$. This captures to a certain degree the level of interpretability of the rule. 
\item \textit{Support} -- the number of observations in the dataset that satisfy the rule, namely $|\{\textbf{x}_i:\left(\prod_{p\in \mathcal{P}'} x_{i,p}\right)=1\}|$. This serves as a measure of coverage for the applicability of the rule. 
\end{itemize}

Note that the fact that the above definitions apply only to rules where features are equal to 1 (and not 0) may seem to be a limitation. However, one can easily extend the feature space by adding binary features that are equal to the complements of the original features, that is, by adding $\textbf{X}^c=1-\textbf{X}$. In this case, rules that contain complement features can be interpreted as rules where an original feature is equal to 0. 
In what follows, we assume that the design matrix $\textbf{X}$ is of dimensions $N\times 2P$ and includes both the original and complement matrices: $[\textbf{X},\textbf{X}^c]$. 

\subsection{Optimization}\label{sub:rules_optimization}

We now describe the formulation of multiple algorithms to generate rules that achieve the objectives of high sparsity and support. 

\paragraph{Optimizing sparsity.}
Let $b_p$ denote a binary decision variable that indicates whether $p\in \mathcal{P}'$.
Denoting $\mathcal{P}_e\overset{\Delta}{=}\{p:x_{e,p}=1\}$ as the largest set of features that ``agree'' with observation $\textbf{x}_e$ allows us to write Condition 1 as $\mathcal{P}'\subseteq \mathcal{P}_e$. Therefore, we need only consider variables $p$ for which $p\in \mathcal{P}_e$. In order for Condition 2 to hold, observations with labels different from $y_e$ must not satisfy the rule $\mathcal{P}'\Rightarrow y_e$. That is, for each such observation $i$, a feature $p\in \mathcal{P}_e$ must be selected for which $x_{i,p}=0$. Each feature $p$ therefore covers a set of observations with opposite labels, and a feasible solution must cover all observations whose labels are different from $y_e$. This is an instance of the \textit{Minimal Set Cover Problem} \citep{feige1998threshold,feige1996threshold}.

More formally, let $A_p\subseteq\{1,\ldots,N\}$ denote the (constant) set of observations $i$ that satisfy $x_{i,p}=0$. Finding a rule with optimal sparsity is the solution to the following optimization problem:

\begin{equation}\label{eq:MaxSparsity}
\begin{array}{ll@{}ll}
\text{minimize}  & \displaystyle\sum\limits_{p\in \mathcal{P}_e} b_{p}      &\\
\text{subject to}& \displaystyle\sum\limits_{p\in \mathcal{P}_e} b_{p} \cdot \mathbbm{1}[\textbf{x}_i\in A_p]   & \geq 1,      &i\in\{i: y^{M}_i\neq y_e\}\\
&b_p \in\{0,1\} & & p\in \mathcal{P}_e.
\end{array}
\end{equation}

We briefly note that we conducted a computational study on the FICO dataset where sparse explanations were generated for each of the 10K observations, based on all other observations. The running time consistently took less than 7 seconds, and the average sparsity was under 3 features. 

\paragraph{Optimizing support.}
Formulation~(\ref{eq:MaxSparsity}) can be extended to incorporate support by adding a binary decision variable $r_i$ (and an appropriate linear constraint) for each observation that indicates that the rule applies to the respective observation. 
An additional constraint was added to limit the number of features to a predefined constant MAX\_SPARSITY in the resulting rule. 

We experimented with our formulations by generating summary-explanations on the FICO dataset. We first solved Formulation~\ref{eq:MaxSparsity}, and set its solution as the value of MAX\_SPARSITY in the modified formulation that optimizes support. We then increased the value of MAX\_SPARSITY by 1 and 2 to relax the respective constraint, in order to improve the support size (at the cost of worse sparsity).

We generated summary-explanations (where we maximized support) for all observations in the FICO dataset. The average number of features in each explanation did not change (comparing with the optimal sparsity) and was equal to 2.9 when MAX\_SPARSITY was set to the solution of Formulation~\ref{eq:MaxSparsity}; sparsity slightly worsened to 3.6 and 4.4 when MAX\_SPARSITY was increased by 1 and 2, respectively. We also found that the number of summary-explanations for which the support is less than 10 was 9.7\% of all observations for the solution given by Formulation~\ref{eq:MaxSparsity}, and was equal to 4.7\%, 1.2\%, and 0.2\% of all observations for the maximal support solution when  MAX\_SPARSITY was increased by 0, 1, and 2, respectively. Clearly, this indicates a tradeoff between support and sparsity; when the constraint on MAX\_SPARSITY is relaxed, the cardinality  of the rule increases and the rule becomes less interpretable, however, at the same time support increases which improves the confidence in the resulting rule. Overall, the average cardinality is nicely low and the support is reasonably high.

\paragraph{Optimization procedure for the challenge.} In an actual deployed system, generating rules could be done offline for each user, since users' credit information changes infrequently over time. In contrast, in our code, we wanted to allow the user the ability to interact with the system to see how explanations are generated for any possible new observation. Therefore, we wanted to limit the running time for generating explanations to provide a reasonably short response. 

To this end, we first created a database of summary-explanations using the FICO dataset. We created summary-explanations for maximal sparsity, and maximal support subject to sparsity constraints of +0, +1, and +2 of optimal. We did the same for 10K additional random observations. When a user clicks to generate an explanation, the database is scanned and the sparsest rule whose support is greater than 10 is returned. Otherwise, if no rules were found, the maximal sparsity optimization problem is solved. If the support of the returned rule is greater than 10, that solution is returned. Otherwise, the maximal support optimization is solved, consecutively relaxing the constraint on the maximal sparsity. If rules with sufficiently large support were not found, the procedure is repeated with minimal support set to 5 (instead of 10). Beyond this, an error message is displayed if no summary-explanations were found, because the observation is an outlier; there is no rule characterizing it.

\begin{figure}
\centering
\includegraphics[width=0.40\linewidth]{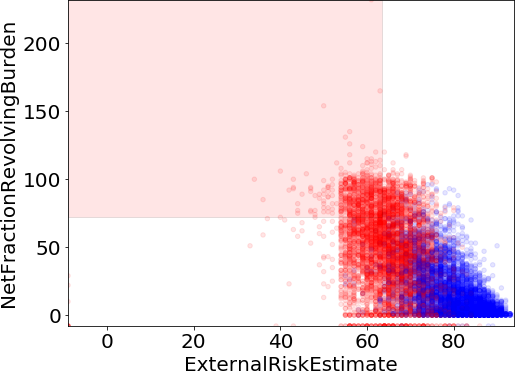}
\qquad
\includegraphics[width=0.54\linewidth]{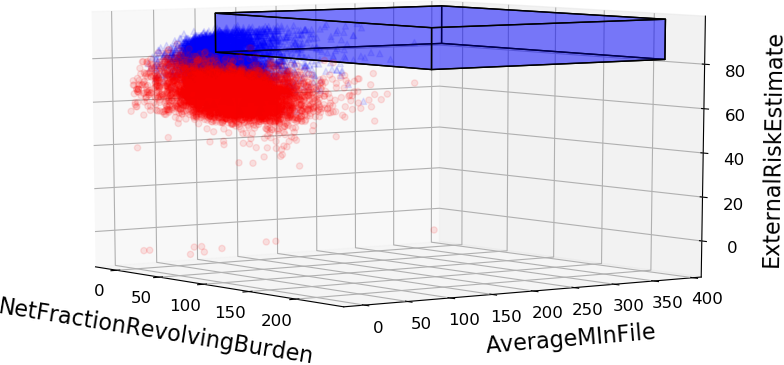}
\caption{\textit{Left:} an explanation for Observation 6 that has support of 700 observations; \textit{Right:} an explanation for Observation 1005 that has support of 990 observations.}
\label{fig:rule2d}
\end{figure}

Figure \ref{fig:rule2d} illustrates the tradeoff between simplicity and coverage for two rules generated by \texttt{SetCoverExplanation}. The rule on the left is 2-dimensional and the rule on the right involves 3 dimensions.

\subsection{Case-based explanations}\label{sub:consistent-cases}

\begin{figure}
  \fbox{\includegraphics[width=\linewidth]{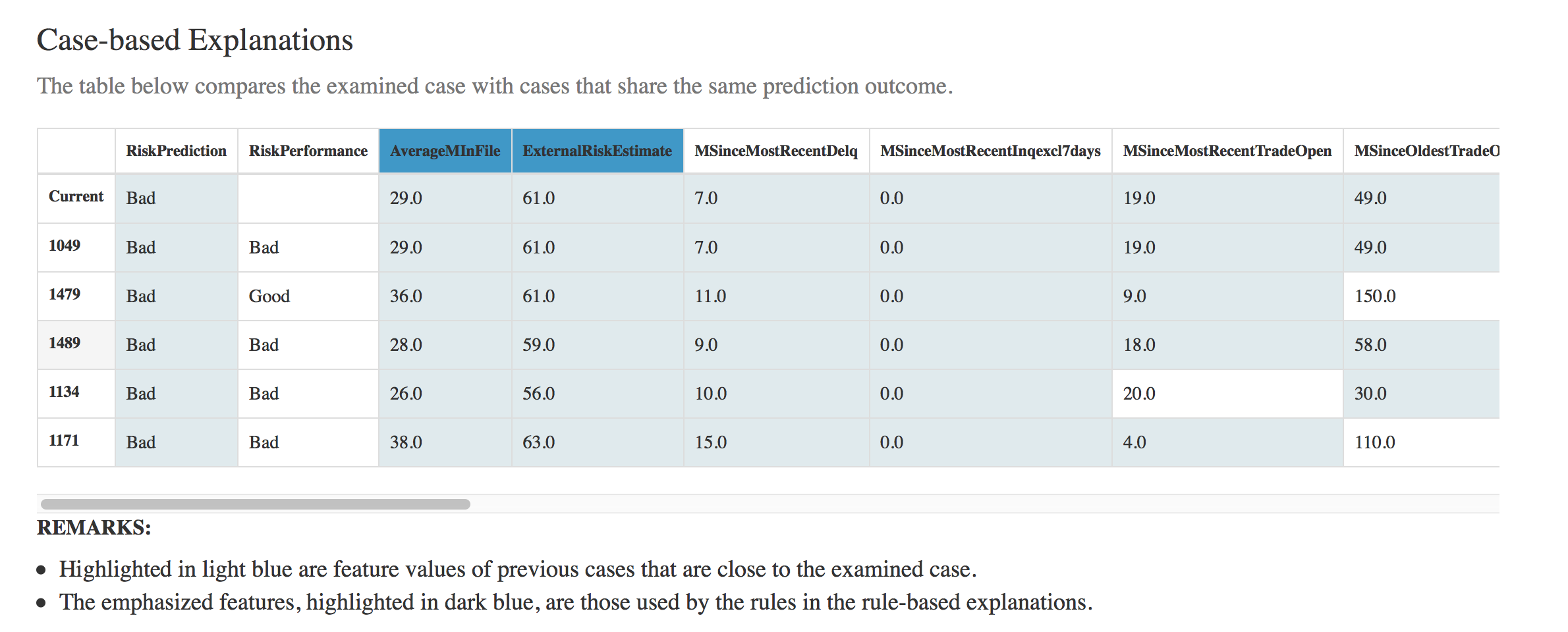}}
  \caption{Case-based explanations.}
  \label{fig:CaseBased}
\end{figure}

Given a previously unseen case, we can identify similar cases from the FICO dataset to assess the prediction made by our global model for the unseen case. To do this, we find all the cases in the dataset that satisfy the consistent rule-based explanation for the unseen case (obtained by \texttt{SetCoverExplanation}). We then rank these cases according to how many binary features (see Section \ref{gen_inst} for how we obtain the binary features) they share with the unseen case, and present the five highest-ranked similar cases to the user. Figure~\ref{fig:CaseBased} gives an example of a case-based explanation: our visualization contains a table showing the current case (top row) and previous cases that are most closely related to the current case (all other rows). 


\section{Thoughts}
\textbf{Future work:} There are several possible extensions to our work. Methods like RiskSLIM \citep{UstunRu2017KDD} could make the subscale scores more interpretable by restricting to integer coefficients. Our visualization interface could be extended to be much fancier, like \citep{Ming2018} for rule-list exploration.

\textbf{Conclusion:}
Since the FICO dataset does not seem to require a black box for good performance, perhaps many other applications in finance also do not require a black box. The answer to this hypothesis remains unclear. However, challenges like the one initiated by FICO can help us to determine the answer to this important question.

\section*{Appendix: Login Information}
The web interface for our system can be found at: \href{http://dukedatasciencefico.cs.duke.edu}{http://dukedatasciencefico.cs.duke.edu} \\
with username dukedatascience and password OxNaUTsSjH0GQ

\bibliographystyle{plainnat}
\bibliography{reference}

\begin{thebibliography}{16}
\providecommand{\natexlab}[1]{#1}
\providecommand{\url}[1]{\texttt{#1}}
\expandafter\ifx\csname urlstyle\endcsname\relax
  \providecommand{\doi}[1]{doi: #1}\else
  \providecommand{\doi}{doi: \begingroup \urlstyle{rm}\Url}\fi

\bibitem[Chen and Rudin(2018)]{chen2018optimization}
Chaofan Chen and Cynthia Rudin.
\newblock An optimization approach to learning falling rule lists.
\newblock In \emph{International Conference on Artificial Intelligence and
  Statistics (AISTATS)}, pages 604--612, 2018.

\bibitem[Feige(1996)]{feige1996threshold}
Uriel Feige.
\newblock A threshold of ln n for approximating set cover (preliminary
  version).
\newblock In \emph{Proceedings of the twenty-eighth annual ACM symposium on
  Theory of computing}, pages 314--318. ACM, 1996.

\bibitem[Feige(1998)]{feige1998threshold}
Uriel Feige.
\newblock A threshold of ln n for approximating set cover.
\newblock \emph{Journal of the ACM (JACM)}, 45\penalty0 (4):\penalty0 634--652,
  1998.

\bibitem[Guidotti et~al.(2018{\natexlab{a}})Guidotti, Monreale, Ruggieri,
  Pedreschi, Turini, and Giannotti]{guidotti2018local}
Riccardo Guidotti, Anna Monreale, Salvatore Ruggieri, Dino Pedreschi, Franco
  Turini, and Fosca Giannotti.
\newblock Local rule-based explanations of black box decision systems.
\newblock \emph{arXiv preprint arXiv:1805.10820}, 2018{\natexlab{a}}.

\bibitem[Guidotti et~al.(2018{\natexlab{b}})Guidotti, Monreale, Ruggieri,
  Turini, Giannotti, and Pedreschi]{guidotti2018survey}
Riccardo Guidotti, Anna Monreale, Salvatore Ruggieri, Franco Turini, Fosca
  Giannotti, and Dino Pedreschi.
\newblock A survey of methods for explaining black box models.
\newblock \emph{ACM Computing Surveys (CSUR)}, 51\penalty0 (5):\penalty0 93,
  2018{\natexlab{b}}.

\bibitem[Lakkaraju et~al.(2017)Lakkaraju, Kamar, Caruana, and
  Leskovec]{lakkaraju2017interpretable}
Himabindu Lakkaraju, Ece Kamar, Rich Caruana, and Jure Leskovec.
\newblock Interpretable \& explorable approximations of black box models.
\newblock \emph{arXiv preprint arXiv:1707.01154}, 2017.

\bibitem[Lou et~al.(2013)Lou, Caruana, Gehrke, and Hooker]{Caruana13}
Yin Lou, Rich Caruana, Johannes Gehrke, and Giles Hooker.
\newblock Accurate intelligible models with pairwise interactions.
\newblock In \emph{Proceedings of the 19th ACM SIGKDD international conference
  on Knowledge discovery and data mining}, pages 623--631. ACM, 2013.

\bibitem[Medical calculators()]{mdcalc}
Medical calculators.
\newblock Mdcalc - medical calculators, equations, algorithms, and scores.
\newblock \url{https://www.mdcalc.com}, 2018.
\newblock Accessed: 2018-11-12.

\bibitem[Ming et~al.(2018)Ming, Qu, and Bertini]{Ming2018}
Y.~Ming, H.~Qu, and E.~Bertini.
\newblock Rulematrix: Visualizing and understanding classifiers with rules.
\newblock \emph{IEEE Transactions on Visualization and Computer Graphics},
  2018.

\bibitem[Moro et~al.(2011)Moro, Laureano, and Cortez]{moro2011data}
S.~Moro, R.~Laureano, and P.~Cortez.
\newblock Using {Data} {Mining} for {Bank} {Direct} {Marketing}: {An}
  {Application} of the {CRISP-DM} {Methodology}.
\newblock In \emph{Proceedings of European Simulation and Modelling Conference
  ({ESM'2011})}, pages 117--121. Eurosis, 2011.

\bibitem[Moro et~al.(2014)Moro, Cortez, and Rita]{moro2014data}
S{\'e}rgio Moro, Paulo Cortez, and Paulo Rita.
\newblock A data-driven approach to predict the success of bank telemarketing.
\newblock \emph{Decision Support Systems}, 62:\penalty0 22--31, 2014.

\bibitem[Regulation(2016)]{regulation2016general}
Protection Regulation.
\newblock General data protection regulation.
\newblock \emph{Official Journal of the European Union}, 59:\penalty0 1--88,
  2016.

\bibitem[Ribeiro et~al.(2018)Ribeiro, Singh, and Guestrin]{ribeiro2018anchors}
Marco~Tulio Ribeiro, Sameer Singh, and Carlos Guestrin.
\newblock Anchors: High-precision model-agnostic explanations.
\newblock In \emph{AAAI Conference on Artificial Intelligence}, 2018.

\bibitem[Shaposhnik and Rudin(2018)]{ShaposhnikRu18}
Yaron Shaposhnik and Cynthia Rudin.
\newblock Globally-consistent rule-based summary-explanations for machine
  learning models, with application to credit-risk evaluation.
\newblock Unpublished, 2018.

\bibitem[Ustun and Rudin(2016)]{ustun2016supersparse}
Berk Ustun and Cynthia Rudin.
\newblock Supersparse linear integer models for optimized medical scoring
  systems.
\newblock \emph{Machine Learning}, 102\penalty0 (3):\penalty0 349--391, 2016.

\bibitem[Ustun and Rudin(2017)]{UstunRu2017KDD}
Berk Ustun and Cynthia Rudin.
\newblock Optimized risk scores.
\newblock In \emph{Proceedings of the 23rd {ACM} {SIGKDD} International
  Conference on Knowledge Discovery and Data Mining}, 2017.

\end{thebibliography}

\end{document}